# Aikyam: A Video Conferencing Utility for Deaf and Dumb

Kshitij Deshpande
*Information Technology*
Pune Institute of Computer Technology
Pune, India
kshitij.deshpande7@gmail.com

Varad Mashalkar
*Information Technology*
Pune Institute of Computer Technology
Pune, India
varadmash2201@gmail.com

Kaustubh Mhaisekar
*Information Technlogy*
Pune Institute of Computer Technology
Pune, India
kaustubh.m0803@gmail.com

Amaan Naikwadi
*Information Technology*
Pune Institute of Computer Technology
Pune, India
amaannaikwadi@gmail.com

Dr. Archana Ghotkar
*Information Technology*
Pune Institute of Computer Technology
Pune, India
aaghotkar@pict.edu

*Abstract* — With the advent of the pandemic, the use of video conferencing platforms as a means of communication has greatly increased and with it, so have the remote opportunities. The deaf and dumb have traditionally faced several issues in communication, but now the effect is felt more severely. This paper proposes an all-encompassing video conferencing utility that can be used with existing video conferencing platforms to address these issues. Appropriate semantically correct sentences are generated from the signer's gestures which would be interpreted by the system. Along with an audio to emit this sentence, the user's feed is also used to annotate the sentence. This can be viewed by all participants, thus aiding smooth communication with all parties involved. This utility utilizes a simple LSTM model for classification of gestures. The sentences are constructed by a t5 based model. In order to achieve the required data flow, a virtual camera is used.

## I. INTRODUCTION

Video Conferencing Platforms (VCPs) have gained popularity in recent times, especially in the wake of the COVID-19 pandemic. While these platforms have made communication easier for most people, they have posed a significant challenge for the deaf and dumb community. As of 2016, 6.3% of the Indian population face significant auditory loss. WHO reports 430 million people worldwide with the need for rehabilitation for hearing loss. Sign Language (SL) is a primary way of communication adopted by the Deaf and Hard of Hearing (DHH) community.

Existing VCPs provide limited support to sign language users and require the need of a third person to act as an intermediary. This research proposes a lightweight desktop-based utility named "Aikyam" which can detect and interpret people communicating in sign language during a video conference call. It is compatible with popular platforms like Google Meet, Teams etc. Using a computationally lightweight algorithm, Aikyam can identify dynamic signs used in ISL and generate the sentence text for what the user is saying in sign language.

The paper targets 2 main aspects of the proposed application in each of the section described. This aspects include the tasks of Sign Language Interpretation and generation of semantic sentences using keywords. The remainder of this paper is structured to cover the following sections. A comprehensive survey of existing methodologies is presented under the literature survey section. The datasets used are discussed in section 3. Section 4 describes the architecture of the utility. The detailed methodology and algorithms used in the utility are explained in section 5. Finally, the sections 6 and 7 discuss the results and conclusion of the research respectively. Section 7 also describes the possible scope in which this research can be expanded.

## II. LITERATURE SURVEY

Interpretation of SL is a progressive area of research which has evolved in various factors such as methods of data acquisition, data representation and classification algorithms. Recognition of gestures is categorized into recognition of characters and numbers, recognition of words and phrases and interpretation of sentences [3]. Real time interpretation of sign language is observed to be done using various permutations and combinations of data acquisition and classification algorithms. 3D sensors like Microsoft Kinect [1 - 5], data gloves [6 - 10] and image processing and computer vision [11 - 18] are observed to be common ways of tracking the user's motion. Xiujuan Chai et al. utilized Kinect sensor to generate a 3D motion trajectory by utilizing the sensor's capability to track depth and color. This trajectory was further subjected to linear resampling to account for the variable speed of the signer. In order to predict the sign, Euclidean distance based calculation was performed between the probe trajectory and the existing database of trajectories representing different signs in the Chinese Sign Language [1].

Computer Vision based approaches also follow a similar structure of feature extraction using image processing techniques attached to learning algorithms for classification of gestures. Techniques and methodologies have been developed to recognize discrete as well as continuous forms of gestures. The features required for classification of data are extracted using techniques like background subtraction [11, 12], body landmarking using deep learning [13, 14], etc. The use of deep learning architectures like Convolutional Neural Networks (CNNs) [11], Long Short Term Memory (LSTM) [13], HMMs [17, 18] and traditional computing algorithms like Dynamic Time Warping (DTW) [3] have been identified as commonly used algorithms for classification of static images or video based data (motion trajectories). One of the research proposed a system where the signer positions his or her gesturing arm in a dedicated section of the window. A background subtraction algorithm extracts the actual hand image which is fed to a 6 layer CNN [11]. HaarCascade classifiers are also used by some researchers in order to classify images received from background subtraction process [12]. A single layer unidirectional LSTM model was utilized by one of the researches that providing a binary output of whether the user is signing or not. For features, landmarks over the user's body were extracted using PoseNet which was further provided to an optical flow calculation algorithm [13]. Combination of CNN with and LSTM was used by a research to classify continuous signs in Chinese Sign Language (CSL). The fully connected layer of the CNN model was further attached to LSTM model for the purpose of continuous data classification [14]. Mediapipe, a body landmark prediction library, combined with a SVM classifier was employed to classify hand-based gestures in one of the research projects [15]. Recurrent Neural Networks preceded by keypoint extraction using OpenPose to detect Korean Sign Language gestures is one of the methodologies adopted by researchers [16].

## III. DATASET

### A. Sign Language Recognition

Various organizations and individuals have recorded datasets for the purpose of standardization, awareness and education about the signs in ISL. Ramakrishna Mission Vivekananda Educational and Research Institute have released a video-based dataset of ISL on an independent platform. It contains a single data record for each word/phrase. The dataset used for training the sign language recognition model has been recorded using traditional web cameras integrated with computers. The dataset consists of 8 signs. In order to classify data over a time series, each data point forms a 2-dimensional vector. A data point in the dataset consists of feature extraction of 30 sequential frames. Let D represent a data point in the dataset and $L^{ij}$ represent the $i^{th}$ feature vector's $j^{th}$ landmark.
D can be represented as

$$D = \begin{bmatrix} L^{11} & \cdots & L^{1n} \\ \vdots & \ddots & \vdots \\ L^{m1} & \cdots & L^{mn} \end{bmatrix} \quad (1)$$

Each row represents a vector of features (landmarks) extracted from the $i^{th}$ frame. 60 samples for each class have been recorded.

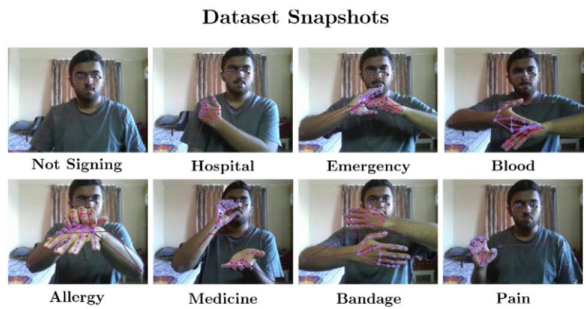

Fig 1. Dataset Instances

### B. Semantic Sentence Generation

This utility currently recognizes signs to construct semantically correct sentences. The dataset contains keywords mapped to the corresponding probable sentences aligned with the healthcare context. In addition to this, the language model is trained using pronouns like he, she, you etc. to ensure more clarity and accuracy during generation. In total, 150 data points have been generated consisting of 1 keyword and permutations of 2 keywords from the set of the valid keywords, which are mapped to corresponding sentences. In order to eliminate the dependency on the order of keywords, the data points consisting permutations of multiple keywords have been mapped to the same sentence irrespective of their order. This way, the algorithm learns the context irrespective of the order of keywords. Following table indicates a subset of the dataset utilized by the sentence generation algorithm.

TABLE I.   SUBSET OF SENTENCE GENERATION DATASET

| Keywords | Sentence |
|---|---|
| allergy | I have an allergic reaction. |
| emergency | It is a medical emergency. |
| allergy, pain | I have an allergic reaction, it is painful |
| pain, allergy | I have an allergic reaction, it is painful |

## IV. ARCHITECTURE

Aikyam primarily aims to facilitate interpretation of sign language and generate meaningful sentences from keywords in commonly available video conferencing (VC) environments. The proposed utility is developed as an installable desktop utility which can be integrated with the VC platforms with changes in settings of the same. The utility is designed to follow a sequential architecture and data passage with appropriate decision making at each stage. The utility is divided into 4 independent layers to introduce logical boundaries, namely:

- User Interface (UI) Layer
- Inferencing Layer
- Virtual Interface Layer
- VC Platforms

The utility provides the user with the following functions:
- Connect/Disconnect Camera
- Start/Stop the interpretation process

The UI layer is responsible for providing a gateway to the users to access the above-mentioned functionalities. It enables ease of access and convenience for the intended users.

### A. Inferencing Layer

Once the user allows the processing of the video feed through the user interface, the inferencing layer is activated. It acts as the decision-making layer of the application. The input video feed is obtained from the camera in the form of sequence of images. This feed is analyzed for predicting the sign performed by the user. The data is stored in a buffer for the classification model to interpret. The buffer is a fixed size queue of sequential data points that represent extracted features from the input image. The buffer is limited to store 30 sequential data points. Let the buffer be represented by B and a data point in the time series be represented as L(i) where L(i) represents a vector of landmarks on i[th] sequence number.

L(i) forms a row vector of extracted features. Here j represents the count of features extracted from each frame.

$$L(i) = [L^{i1}, L^{i2}, L^{i3} \dots, L^{ij}] \qquad (2)$$

Finally, B consists of a sequential combination of L(i) vectors given as the transpose of the following matrix

$$B = [L(1), L(2), L(3) \dots, L(30)]^T \qquad (3)$$

The structure of B is similar to that of a data point in the dataset represented in equation 1. Once the buffer is full, the classification process can be started. The classification of the performed gesture is further followed by the process of generating meaningful sentences from the signs detected so far. This is subject to whether the user requests for the generation of the sentence manually through the UI. The detected signs are stored in a separate data structure that maintains the sequence in which the signs were detected and ensures non-duplication of signs in the buffer. The user can signal the application to generate a meaningful sentence from the detected signs through the user interface. An NLG model is used to convert the detected keywords into semantically correct sentence sensitive to the desired context. The sentence is annotated to the user's feed and displayed directly on the VC platforms. Parallelly, the text to speech module converts the sentence into an audio format and plays the generated audio through the virtual microphone. In this way, the audience of the conference can actually listen to what the signer needs to communicate.

*B. Virtual Interface Layer*

The Virtual Interface Layer is responsible for communicating the results derived from the inferencing layer to the VCP. This mainly requires rerouting and control over the audio and video transmission of data to the VCP. The utility is primarily designed to bypass the traditional mechanisms of video and audio transmission and reroute the same to the VCP through the application for introducing an additional layer of processing. Hence, the virtual interface layer helps to tap into traditional data flow, process the data accordingly and retransmit the processed data to the VCP.

The VCP layer is responsible for handling the traditional video conferencing needs and is managed by the respective platforms (Google Meet, Zoom, Microsoft Teams, etc.). Fig. 2 represents the architecture of the proposed system explained above. It indicates the structure and interaction of the layers within the utility.

V. ALGORITHMS

*A. Feature Extraction*

Feature extraction marks the initial step of the video processing component of the utility. Various ways have been identified in the survey for acquiring meaningful data from the input images or videos. Techniques such as background subtraction [11, 12] and body landmarking [13, 14] are popularly employed. Background subtraction mainly requires the user to perform the sign in a designated area of the viewport. This introduces a restriction on the signer thereby reducing the practicality in the process. Sensors, both on and off the user's body, can be used to capture physical landmarks on the user's body. Such sensors usually utilize variations in current due to change in resistance or depth mapping using infrared sensors. With recent developments in computer vision and deep learning techniques, image-based extraction of body landmarks has seen substantial improvement in terms of accuracy and real time performance. Pose estimation capabilities of MediaPipe's heuristic model allows extraction of 543 landmarks in totality, split into the following landmark categories

- 33 Pose Landmarks
- 468 Face Landmarks
- 21 Hand Landmarks per hand

The utility currently focusses on the classification of motion of hands and other body parts. Hence the facial landmarks are discarded and the rest of the pose landmarks along with the hand landmarks are retained. The library returns the landmarks in the form of x and y coordinates in a vector. Out of the 543 landmarks, only 129 points are utilized by the utility. One single data point (L(i)) is a single row vector as mentioned in equation 2. The structure L(i) consists of these points with alternating x and y coordinates, it can be given as mentioned in the following equation.

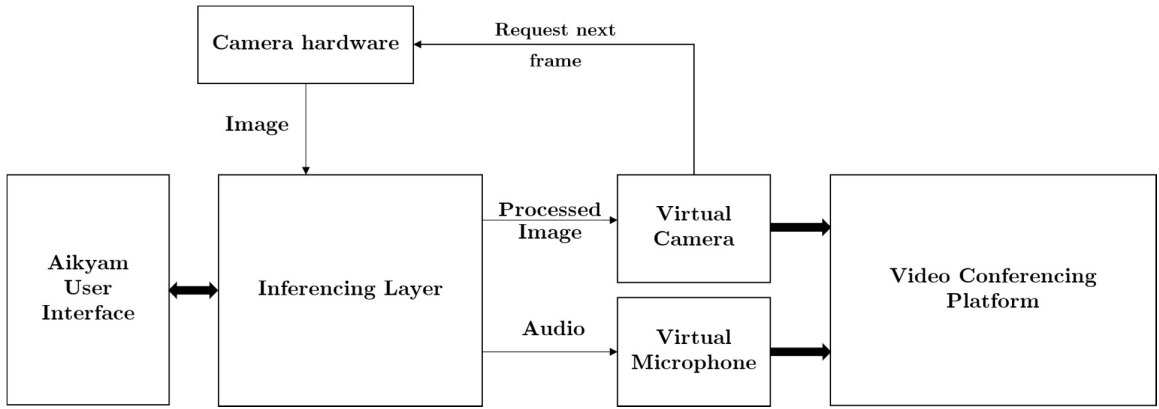

Fig 2. Layered Architecture

$$L(i) = [x^1, y^1, x^2, y^2, ..., x^{129}, y^{129}] \qquad (4)$$

This forms the feature vector derived from one single frame. As mentioned in equation 3, these feature vectors are stacked to form the input for the classification network. The details of the network are mentioned in the following section. The following image shows the points detected by MediaPipe on the human body. These points consist of landmarks on the face, shoulder, elbow, palm, waist and the knees.

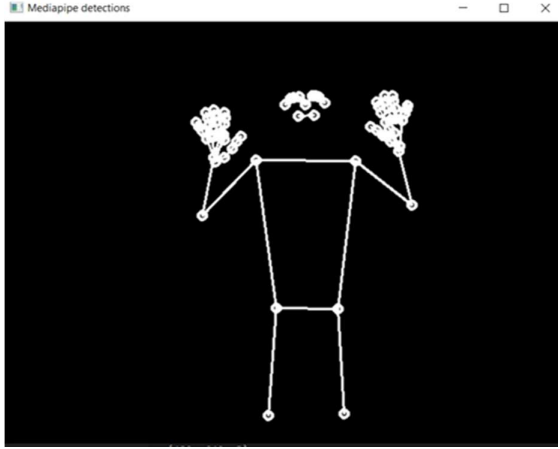

Fig 3. Landmarks detected by MediaPipe on human body

*B. Sign Language Interpretation*

Mediapipe is used to generate a vector which consists of landmarks on the body. This forms the feature vector derived from one single frame. Whilst processing the video frames, a fixed size buffer is maintained to keep track of the user's motion in the form of feature vectors. This buffer is passed to the model for predicting the gesture. A separate counter is maintained to keep track of the predictions over time. The most frequent prediction is considered as the final output. This counter-based mechanism helps overcome the limitations of the model in real time processing. The counter coupled with the interpretation model allow for a more stable prediction by dealing with fluctuations in output.

**Sign Language Interpretation Algorithm**

1:  landmarkBuffer ← []
2:  maxBufferSize ← 30
3:  counter ← {}
4:  **while** true **do**
5:      frame ← camera
6:      landmarkVector ← frame
7:      landmarkBuffer ← landmarkVector
8:      **if** landmarkBuffer.size > maxBufferSize **then**
9:          landmarkBuffer.dequeue()
10:         result ← model.classify(landmarkBuffer)
11:         counter[result] ← counter[result] + 1
12:     **end if**
13: **end while**

The architecture of the model can be represented as follows.

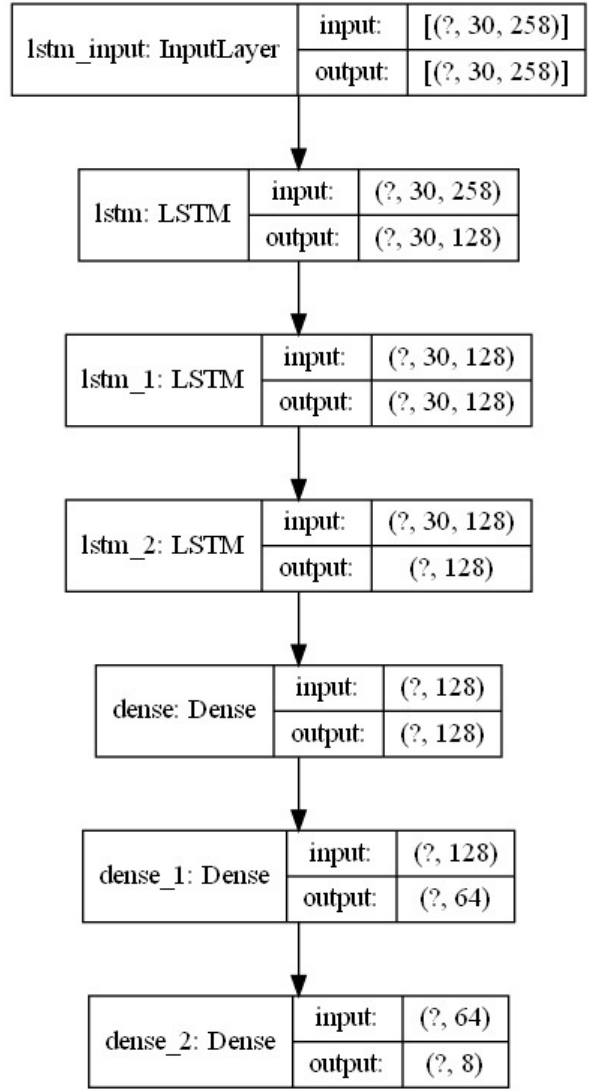

Fig 4. Architecture of Sign Language Interpretation Model

*C. Video Segmentation*

The signs performed by the user are of continuous and sequential nature. Separation of one sign from another becomes a concern to obtain accurate results from the classification algorithm. The classification algorithm is trained to detect whether the signer is performing a sign or is idle. Whenever the user is signing, the frequency counter is incremented for the predicted sign. The signer needs to wait for a fixed amount of time (predetermined threshold) before performing the next sign. If the segmentation algorithm detects that the user is not signing for more than the threshold time, then the algorithm picks the most occurring sign from the frequency counter. This sign is added to a sequence of keywords which is then used to generate a sentence. The timer and frequency counter are reset for capturing the next sign.

The algorithm for sign segmentation can be stated as follows–

**Sign Segmentation Algorithm**

```
1: keywordBuffer ← []
2: timer ← 0
3: timeThreshold ← 5
4: while true do
5:     result ← model.classify(landmarkBuffer)
6:     if result = "notsigning" then
7:         if timerisreset then
8:             timer.start()
9:         else
10:            if timer.elapsedTime() >= timeThreshold then
11:                if resultnotinkeywordBuffer then
12:                    keywordBuffer.append(result)
13:                end if
14:                timer.reset()
15:                counter.reset()
16:            end if
17:        end if
18:    else counter[result] ← counter[result] + 1
19:    end if
20: end while
```

## VI. RESULTS

This section describes the results obtained from the algorithms and the utility in totality. The computed metrics for the classification algorithm and the output of the sentence generation algorithms are discussed. Computation and inference time involved in generating the output have also been stated.

### A. Sign Language Interpretation Model Performance

The LSTM network is able to classify the eight-sign vocabulary at an overall accuracy of 96.45%. The model was trained for 200 epochs, observing a gradual increase in accuracy. Parallelly, the use of DTW method coupled with kernel Nearest Neighbor(kNN) algorithm was also explored in order to account for the factor of variable speed of gesturing. The DTW-kNN algorithm compares the current data point with the existing data points in the dataset and uses DTW as a distance metric to provide the output using kNN's traditional workflow. This algorithm was found to be more accurate at the cost of latency. The comparative analysis of LSTM network and the DTW-kNN algorithm is presented in the following data.

TABLE II. COMPARITIVE ANALYSIS OF DTW-KNN ALGORITHM AGAINST LSTM NETWORK

| Metric | DTW-kNN | LSTM Model |
|---|---|---|
| Training Accuracy | 100% | 96.45% |
| Average Prediction Time | >1 minute | 0.0866 seconds |

For real time prediction, it is crucial to have a faster algorithm. Hence, the LSTM network is used to classify the gestures. Since the dataset for the model was generated manually, the entire dataset was utilized for training purpose. The tests on the model were carried out by placing manual subjects in front of the camera and recording multiple instances for the different considered signs. These instances were subjected to classification and the result was formulated on the same. The details of the classification performed by the LSTM model on testing data can be observed in the following table. The model was able to correctly predict 91.66% of the testing set.

TABLE III. CLASSIFICATION DETAILS OF LSTM MODEL ON TESTING DATA

| Class | Total Samples | True Positives | False Positives |
|---|---|---|---|
| Not Signing | 30 | 27 | 5 |
| Blood | 30 | 26 | 1 |
| Medicine | 30 | 25 | 0 |
| Allergy | 30 | 30 | 3 |
| Emergency | 30 | 26 | 5 |
| Hospital | 30 | 29 | 3 |
| Bandage | 30 | 27 | 0 |
| Pain | 30 | 30 | 3 |

### B. Sentence generation

After fine-tuning the T5 model with keywords from the healthcare domain, it was observed that the performance improved significantly.

TABLE IV. OUTPUT COMPARISON OF SENTENCE GENERATION MODEL

| Keywords | Expected Output | Actual Output |
|---|---|---|
| medicine, pain | I am in pain, what medicine should I use? | I am in pain, what medicine should I use? |
| emergency, allergy | I have an allergic reaction, it is an emergency | I have an allergic reaction, it is an emergency |
| cough | I have a cough | I have a cough |
| hospital, cold | I have a cold, should I rush to the hospital | I have a cold, should I rush to the hospital |
| sick, accident | I had an accident, I am feeling sick | I have an accident, it is sick |

## VII. CONCLUSION AND FUTURE WORK

The proposed utility in this paper wishes to bridge the communication gap between the DHH community and the hearing individuals in the video conferencing environment by processing real time data. The interpretation of sign language was approached by 2 different techniques, the result of which indicated that use of deep learning models, more specifically neural networks with the capacity to analyze time series data prove to be a good choice to deal with dynamic gestures. The survey conducted for such techniques also supports this fact. The classification of continuously performed signs is one of the key issues according to the survey. The key advantage of the algorithms proposed for identifying sequentially performed signs is that with the introduction of the frequency counter, the model can make stable predictions.

The sentence generation algorithm provides an additional context to the conversation by improving the quality of communication. A common user may not understand the context just on the basis of the keywords identified by the interpretation model. The interpretation model was able to give 91.66% accuracy on the testing set of data. Another major advantage of this utility as a software is that with a comparatively lower resource footprint, the utility can be targeted towards lower end devices and can be made available to a larger part of the community.

Considering the future aspects of this utility, there are multiple ways in which this architecture and algorithms can be taken forward. For instance, the primary objective of further development will begin with the scaling of interpretation model to interpret a larger set of ISL vocabulary. Parallelly, the sentence generation model will require additional context to generate more meaningful sentences. The current architecture is implemented on computers. With the increased use of smartphones, the utility can also be developed for such handheld devices by changes in the architecture. Clear separations of concerns with respect to the interfacing and inferencing into the client server paradigm can be the best possible way to increase the scope of the utility.

REFERENCES

[1] Chai, Xiujuan, et al. "Sign language recognition and translation with kinect." IEEE conf. on AFGR. Vol. 655. 2013.

[2] Verma, Harsh Vardhan, Eshan Aggarwal, and Satish Chandra. "Gesture recognition using kinect for sign language translation." 2013 IEEE Second International Conference on Image Information Processing (ICIIP-2013). IEEE, 2013.

[3] Ghotkar, Archana S., and Gajanan K. Kharate. "Dynamic hand gesture recognition and novel sentence interpretation algorithm for Indian Sign Language using microsoft kinect sensor." Journal of Pattern Recognition Research 1 (2015): 24-38.

[4] Hazari, Shihab Shahriar, Lamia Alam, and Nasim Al Goni. "Designing a sign language translation system using kinect motion sensor device." 2017 International Conference on Electrical, Computer and Communication Engineering (ECCE). IEEE, 2017.

[5] Li, Kin Fun, Kylee Lothrop, Ethan Gill, and Stephen Lau, "A web-based sign language translator using 3d video processing." 2011 14th International Conference on Network-Based Information Systems. IEEE, 2011.

[6] Abhishek, Kalpattu S., Lee Chun Fai Qubeley, and Derek Ho. "Glove-based hand gesture recognition sign language translator using capacitive touch sensor." 2016 IEEE international conference on electron devices and solid-state circuits (EDSSC). IEEE, 2016.

[7] Das, Abhinandan, et al. "Smart glove for sign language communications." 2016 international conference on accessibility to digital world (ICADW). IEEE, 2016.

[8] Alzubaidi, Mohammad A., Mwaffaq Otoom, and Areen M. Abu Rwaq. "A novel assistive glove to convert arabic sign language into speech." ACM Transactions on Asian and Low-Resource Language Information Processing 22.2 (2023): 1-16.

[9] Praveen, Nikhita, Naveen Karanth, and M. S. Megha. "Sign language interpreter using a smart glove." 2014 international conference on advances in electronics computers and communications. IEEE, 2014.

[10] Shukor, A. Z., Miskon, M. F., Jamaluddin, M. H., Ibrahim, F. bin A., Asyraf, M. F., & Bahar, M. B. bin (2015). A New Data Glove Approach for Malaysian Sign Language Detection. Procedia Computer Science, 76, 60-67. https://doi.org/10.1016/j.procs.2015.12.276

[11] R. Harini, R. Janani, S. Keerthana, S. Madhubala and S. Venkatasubramanian, "Sign Language Translation," 2020 6th International Conference on Advanced Computing and Communication Systems (ICACCS), Coimbatore, India, 2020, pp. 883-886, doi: 10.1109 /ICACCS48705.2020.9074370.

[12] K. Dabre and S. Dholay, "Machine learning model for sign language interpretation using webcam images," 2014 International Conference on Circuits, Systems, Communication and Information Technology Applications (CSCITA), Mumbai, India, 2014, pp. 317-321, doi: 10.1109/CSCITA.2014.6839279.

[13] A. Moryossef, I. Tsochantaridis, R. Aharoni, S. Ebling, and S. Narayanan, "Real-Time Sign Language Detection using Human Pose Estimation," arXiv:2008.04637 [cs.CV], 2020.

[14] 14 17. Yang, Su, and Qing Zhu. "Continuous Chinese sign language recognition with CNN-LSTM." Ninth international conference on digital image processing (ICDIP 2017). Vol. 10420. SPIE, 2017.

[15] Halder, Arpita, and Akshit Tayade. "Real-time vernacular sign language recognition using mediapipe and machine learning." Journal homepage: www. ijrpr. com ISSN 2582 (2021): 7421.

[16] Ko, Sang-Ki, H. "Neural sign language translation based on human keypoint estimation." Applied sciences 9.13 (2019): 2683.

[17] Starner, Thad, and Alex Pentland. "Real-time american sign language recognition from video using hidden markov models." Proceedings of International Symposium on Computer Vision-ISCV. IEEE, 1995.

[18] Grobel, Kirsti, and Marcell Assan. "Isolated sign language recognition using hidden Markov models." 1997 IEEE International Conference on Systems, Man, and Cybernetics. Computational Cybernetics and Simulation. Vol. 1. IEEE, 1997.